\documentclass[letterpaper, 10 pt, conference]{ieeeconf}  

\IEEEoverridecommandlockouts                              

\usepackage{amsmath} 
\usepackage{amssymb}  
\usepackage{balance}
\usepackage{xspace}
\usepackage{csquotes}
\usepackage{hyperref}
\usepackage{stmaryrd}
\usepackage{cite}
\usepackage[svgnames,dvipsnames]{xcolor}
\definecolor{shadeGray}{rgb}{0.9,0.95,0.95}
\usepackage[bordercolor=red, backgroundcolor=shadeGray, linecolor=red, textwidth=0.8in, textsize=footnotesize]{todonotes}
\usepackage{listings}
\usepackage{tikz}
\usepackage{pgfplots, pgfplotstable}
\usepackage{subfig}
\usepackage{paralist}
\usepackage{subfig}
\lstdefinestyle{corelang}{
  rulecolor=\color{gray},
  morekeywords={
  repeat, times, foreach, visit, pick, drop, containing, avoiding, while, strict,
  move, right, move, left, in, every, is, color, shape, possible, if, at, and, or, minus, not
  },
  keywordstyle=[2]\emph,
  keywords=[2]{world, point, robot, item, red, blue, yellow, green, triangle, square, circle},
  sensitive=true,
  morecomment=[l]{//},
  morecomment=[s]{/*}{*/},
  commentstyle=\color{gray},
  showstringspaces=false,
  mathescape=true,
  numberstyle=\tiny,
  basicstyle=\footnotesize\ttfamily,
  numbersep=5pt,
  stepnumber=2,
  numbers=none,                   
  morestring=[b]"
}
\usepackage[T1]{fontenc}  
\usepackage[scaled=0.80]{beramono}  
\allowdisplaybreaks

\usepackage{amsthm}
\usepackage{thmtools}

\newcommand{\tool}{Flipper\xspace}

\newcommand{\until}[2]{#1\; \mathcal{U}\; #2}

\title{\LARGE \bf Precise but Natural Specifications for Robot Tasks
}

\author{Ivan Gavran$^1$  \quad Brendon Boldt$^2$\quad Eva Darulova$^1$ \quad Rupak Majumdar$^1$%
\thanks{
$^1$Max Planck Institute for Software Systems, Germany,
$^2$Marist College, USA
}%
\thanks{Brendon Boldt was supported by a DAAD RISE Internship.}
}

\begin{document}

\maketitle
\thispagestyle{empty}
\pagestyle{empty}

\begin{abstract}
We present \tool, a
natural language interface for describing
high level task specifications for robots
that are compiled into robot actions.
\tool starts with a formal core language for task planning that allows
expressing rich temporal specifications and
uses a semantic parser to provide a natural language interface.
\tool provides immediate visual feedback by executing an automatically
constructed plan of the task in a graphical user interface.
This allows the user to resolve potentially ambiguous interpretations.
\tool extends itself via \emph{naturalization}: its users can
add definitions for utterances, from which \tool induces new rules and adds them to the core language,
gradually growing a more and more natural task specification language.
\tool improves the naturalization by generalizing the definition provided by users.
Unlike other task-specification systems, \tool enables natural language
interactions while maintaining the expressive power and formal precision of a programming language.
We show through an initial user study that natural language interactions and generalization
can considerably ease the description of tasks.
Moreover, over time, users employ more and more concepts outside of the initial core language.
Such extensions are available to the \tool community, and users can use concepts that others have defined.

\end{abstract}


\section{Introduction}

As robots move from controlled factory environments to homes, an important challenge is to allow end users to
specify tasks for the robot in clear and unambiguous ways.
%
%
While there are many programming languages for specifying tasks~\cite{golog, strips, pddl} and a number of tools that
compile task specifications to robot action plans~\cite{fainekosTemporalLogicMotionPlanning,hoffmanFF,ankushDrona},
their use is limited to users with programming experience and who have mastered the syntax and semantics of the particular language.

In order to make these languages accessible to end users, one can add ``syntactic sugar'' to a programming language.
In its simplest form, users can express commands in a fixed structured subset of natural language~\cite{hadasTranslatingStructuredEnglish};
utterances beyond this subset are rejected.
Alternatively, one can use rich language models developed in the NLP community for arbitrary dialog understanding.
However, the application to robot task planning is usually limited to a fixed set of the most common 
scenarios~\cite{hadasProvablyCorrectReactiveControlFromNaturalLanguage,thomasonDialog,kollarDialog};
utterances beyond the pre-programmed tasks are rejected. 
Thus, users are either limited to a handful of restricted natual language idioms, 
or they can use their language freely, but the robot can only understand a few actions.

In this paper, we present \tool, a system that aims to keep the precision of a programming language while allowing the ease of use of natural language.
It does so by interactively extending the underlying grammar of a core formal language through induction of general rules from definitions.
\tool is able to adapt to the language style of its users while always maintaining a connection to the underlying programming language.

The core of \tool is a high-level formal programming language for task specification and an executor that transforms that language into a robot action plan.
A \emph{semantic parser} \cite{berantSempre} translates natural language utterances into this core language, 
allowing more flexibility in specifying tasks.
While the core language is always available to programmers,
\tool further allows to extend the capability of the language and parsing through a
process of \emph{naturalization}~\cite{wangVoxelurn}.
Whenever the semantic parser fails to parse an utterance, the user can \emph{define} the
new utterance in terms of a sequence of utterances already understood (i.e., parsable) by the system.
\tool induces a set of new production rules in the core language 
from the user's definition, thus extending the language.
Furthermore, because the user's definition might be too specific (i.e.\ work only for the current state of the world), 
\tool uses semantic-preserving rewritings to generalize to an equivalent definition, matching the utterance and language model better.

Thus, over time, \tool builds up a large lexicon of defined concepts through user interaction and generalization.
Initial users program in the core language, but build up idioms natural to the domain; future users
can use all concepts previously defined by the community.
Along the way, each parsed form retains the precise semantics of the core language.
In that way, \tool helps bridge the gap between expressive and precise programming languages with restricted syntax on one side,
and flexible but ambiguous natural language on the other.

How is \tool 's naturalization mechanism different from defining concepts, for example using the library mechanism, in a programming
language? 
\tool exploits two key features of the robotics domain.
First, it can provide quick visual feedback to the user about the \emph{effect} of an utterance by simulating the plan on the world.
This allows quick resolution of ambiguity.
Second, unlike a library in a general purpose language in which a programmer parameterizes a function explicitly, \tool uses grammar induction to create new parsing rules from definitions.
Finally, it improves users' definitions using an additional generalization mechanism based on rewritings.
While grammar induction and rewriting may not be powerful enough for general-purpose computation, it works remarkably well
in the restricted context of planning worlds.

\begin{figure}[t]
\resizebox{\linewidth}{!}{
\includegraphics{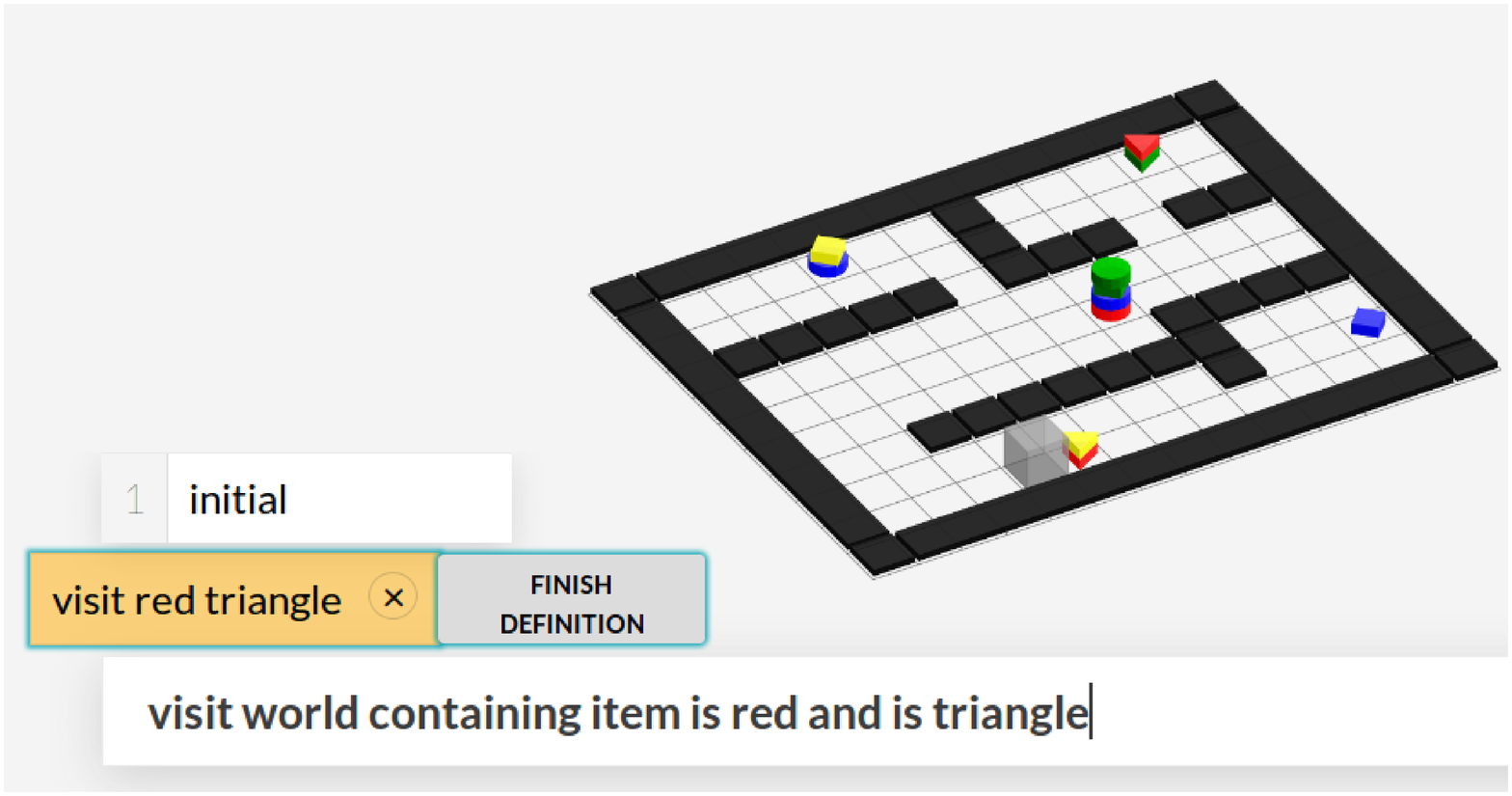}
}
\caption{Simulation world of \tool}
\label{fig:simulationWorld}
\end{figure}

\smallskip
\noindent
\textbf{Interacting with \tool}
The user of \tool instructs a single robot which moves in a world
consisting of several connected rooms separated by walls, presenting obstacles for the robot.
These rooms contain a number of items with different properties; for simplicity we consider here different
colors and shapes.
The robot can modify this world by moving to free locations,
and by applying a set of actions, such as picking up or dropping items.
Figure~\ref{fig:simulationWorld} shows a simulation of such a world (a robot symbol is a grey cube).
The user specifies actions for the robot as an utterance.
For example, the user might ask the robot to
\emph{visit red triangle}.
If \tool cannot understand an utterance, it will ask the user to define this utterance in terms of one or more commands 
that it already understands.
In this example, the user
can define the utterance in the core language: \lstinline{visit world containing item is red and is triangle}.
Based on that definition, \tool induces new grammar rules which allows it to also understand related utterances such as ``visit yellow circle.''

Once \tool successfully parses an utterance into a task specification, it constructs a plan over the simulated
world, and provides visual feedback first, showing what the robot would do.
This is especially useful if there are multiple possible interpretations, e.g., due to ambiguities in the extended grammar.
Once the user selects an interpretation, the robot proceeds with an action (ideally, in the real world; in our current implementation, still in the simulation). 
In case the plan is unrealizable, the user receives feedback specifying which primitive part of the
action is not realizable.

\tool integrates research performed in several different communities---NLP, planning, and program synthesis---in a non-trivial
way.
Overall, it provides an intuitive interface for end-user programming of robot tasks. 
The combination enables non-programmers to interact with robots in a flexible syntax, while assuming a number of power-users 
capable of grounding each utterance to a formal language.
Specifically, we present the following contributions:
\begin{itemize}
	\item an implementation of \tool, a natural language interface for communicating with robots using naturalization, a process of interactive extending of a core programming language,
	\item  improvement to the naturalization process using generalization by semantic-preserving rewriting, 
	\item a preliminary study on 31 users showing that naturalization
helps specify tasks in a more succinct way, that users define derived concepts as building blocks,
and these derived concepts are reused by later users as their basis for interacting with the system.
\end{itemize}

\section{Related work}

There are many systems commanding or communicating naturally with robots as well
as creating programming languages for robots usable by non-programmers, starting
with the seminal work in SHRDLU~\cite{shrdlu}, and continuing over the
years~\cite{kollarDialog,thomasonDialog,roboFlow}.

Formal and logical languages for planning have a long research tradition in AI
and formal methods \cite{golog,pddl,strips,hadasTranslatingStructuredEnglish}. 
Recently, there has been a focus on translating
natural language into formal task specifications~\cite{hadasTranslatingStructuredEnglish,hadasLTLMop,
hadasProvablyCorrectReactiveControlFromNaturalLanguage} to bridge the gap between
end users and formal languages.
These approaches either use a general-purpose NLP pipeline or a fixed set of ``structured'' templates.
For example,~\cite{hadasProvablyCorrectReactiveControlFromNaturalLanguage} uses an NLP pipeline 
but restricts  the set of actions a robot can perform to a fixed set of pre-programmed behaviors.
\cite{hadasTranslatingStructuredEnglish} uses structured templates for a rich class of linear temporal
logic specifications, but the user must only use the templates available.
In contrast, \tool allows both a natural style free of particular templates and a rich class of specifications.
However, it assumes that the process of grammar extension is aided by users who are able to define
natural language utterances in terms of previous utterances and ultimately the programming language. 

The problem of grounding utterances to spatial relations between objects (but without naturalization)
is tackled in~\cite{tellexGrounding,paulGrounding}  by introducing a hierarchical structure
that connects expressions such as \emph{beside the truck} and \emph{beside the box}. 
These concepts can be put in the context of reactive temporal commands~\cite{boteanuVerifiableGrounding}.
It will be interesting to extend \tool with reasoning about spatial relations in this way.

\tool is inspired by and builds upon the work on naturalization of formal
languages in Voxelurn~\cite{wangVoxelurn}, which considers a block world
where a user can build various shapes of different colors. The application of
naturalization to a robot world introduces new challenges: the language contains
declarative and unrealizable commands and dynamic behavior that changes the
state of the world. 
Additionally, \tool improves naturalization by generalizing user-supplied definitions
by program rewriting, which gives the power of defining new concepts to less programming-savvy users.

A similar approach of learning the language from users is
presented in~\cite{azariaLia}, but in the context of personal assistants.
User's feedback is employed in~\cite{iyerLearningNeuralSemanticParser} to
minimize the effort needed for additional annotation of data and iteratively
improve their semantic parser that translates natural language utterances to SQL
queries. 

Beyond the robotics domain, an ensemble of a neural network and logistic regression models is used in~\cite{linTelina,beltagyIFTT} to translate a task 
description from programming-help websites or IFTTT datasets into executable scripts. 
This line of work enables semantic parsing from less direct
instructions, but is not easily adaptable to users interactively giving clues to
the system about the meaning of the utterance. 
Also, these systems require lots of training data to function, whereas \tool extends itself gradually.

\section{Background: Semantic Parsing with Naturalization}
\label{sec:voxelurn}

\tool builds upon the idea of interactive semantic parsing, 
first presented in Voxelurn~\cite{wangVoxelurn}.
In this section we describe the main components of an interactive semantic parser,
using examples from the \tool domain. 

\tool is based on a core language whose  
syntax is defined in~\autoref{fig:core-syntax}.
We shall describe the syntax in detail later; our simple
examples should be understandable without a detailed understanding of the syntax.


We use \emph{semantic parsing}~\cite{liangSemanticParsers}
to parse utterances.
Semantic parsing converts a natural language utterance into
a ranked list of abstract syntax trees. 
%
The semantic parser outputs a ranked list because an utterance might be ambiguous, 
i.e. it could be parsed in several different ways.
To rank the potential parses, the parser uses a statistical model $p_\theta(d | x, u)$ which
assigns probabilities to derivations $d$, given utterance $x$ by user
$u$ and using a set of parameters $\theta$.
The features include whether the rule comes from the core language or is induced, whether
the author of the rule is the same person as $u$, etc.
The user visually inspects (in simulation) the three best-ranked derived programs.
Based on the user's choice, the model's parameters $\theta$ are updated.


If the semantic parser is unable to parse an utterance, it asks the user to
provide a definition for the utterance. 
A grammar induction module takes the definition for a particular utterance and creates one or more new rules
in the underlying grammar for the language. 
The new rules extend the underlying language: in the future, the utterance as well as its variants become parsable.

We will use symbol $x$ for utterance, and $y$ for the provided definition.
While $y$ must be fully parsable using the current grammar rules, only some
parts of $x$ may be parsable. 
In order to induce new rules, the system identifies
\textit{matches}---parsable spans appearing in both $x$ and $y$. 
A set of non-overlapping matches is called a \emph{packing}, and is the
basis for generating new grammar rules.
New grammar rules are introduced through
\emph{simple packing}, \emph{best packing}, and \emph{alignment} \cite{wangVoxelurn},
which we describe below.

We illustrate simple and best packing on the following example.
Suppose a user writes the  (yet underfined) command $x:$  \lstinline{pick 3 items}. 
\tool asks for a definition and the user responds with $y:$ \lstinline{repeat 3 times pick item}.

Simple packing considers pre-defined primitive categories for matching (such as colors, shapes or numbers).
The only primitive match in the above example is the number 3, which in Flipper's grammar has category \lstinline{Num}. 
Therefore, a new rule is added to the grammar: \lstinline{Act $\to$ pick item Num $::=$ repeat Num times pick item}.

Best packing considers maximal packings, i.e. those that would become overlapping by adding any other derivation,
and chooses the packing that scores the best under the model $p_\theta$. 
The best scoring maximal packing for our example results in the rule
\lstinline{Act $\to$ ItemAct Num items$::= $ repeat Num times ItemAct item}. Note that this rule is more
general than the one generated from the simple packing; 
with this rule in the grammar, \tool will in the future understand commands such as \lstinline{drop 2 items}.
(While best packing is more general, it can potentially result in incorrect rules added to the grammar.)

\emph{Alignment} is the third way to induce new rules and it considers the case when the utterance $x$ and the derivation $y$ align almost perfectly.
As an example, for \lstinline{$x:$ throw item} and \lstinline{$y:$ drop item}, a new rule is added
 \lstinline{ItemAct $\to$ throw $::=$ drop}.

%


\section{Flipper}
\label{sec:flipper}

In this section we describe the Flipper language design, implementation
and the generalization technique for user-supplied definitions. \tool and its implementation are publicly available at \url{flipper.mpi-sws.org}.
\subsection{Core Language}
  \tool{}'s \emph{core language} allows users to specify temporal tasks for a
  robot in an abstract world. \autoref{fig:core-syntax} shows the most
  interesting subset of the syntax of the core language.
  While the abstract world in which our robot operates may seem relatively simple,
  the core language allows expressing many interesting scenarios.
  \begin{figure}
    \begin{lstlisting}
// Control flow
Stmt $\to$ Act $\mid$ Stmt; Stmt $\mid$ repeat Num times Stmt
      $\mid$ foreach point in Area Stmt $\mid$ if Cnd Stmt $\mid$ while Cnd Stmt

// Actions:
Act     $\to$ visit Area $\mid$ visit Area while avoiding Area
         $\mid$ move right $\mid$ move left $\mid$ $\ldots$
         $\mid$ ItemAct QItm $\mid$ strict Act
ItemAct $\to$ pick $\mid$ drop

// Locations:
Area $\to$ world $\mid$ Pnt $\mid$ [Pnt, Pnt, ..., Pnt] $\mid$ Area containing Itm
     $\mid$ Area and Area $\mid$ Area or Area $\mid$ Area minus Area
Pnt  $\to$ [Num, Num] $\mid$ point

// Items:
QItm $\to$ every Itm $\mid$ Itm
Itm  $\to$ item $\mid$ item Fltr
Fltr $\to$ is Prop $\mid$ Fltr and Fltr $\mid$ Fltr or Fltr $\mid$ not Fltr
Prop $\to$  C $\mid$  S
C    $\to$ red $\mid$ blue $\mid$ green $\mid$ yellow $\mid$ $\ldots$
S    $\to$ triangle $\mid$ square $\mid$ circle $\mid$ $\ldots$
  
//Conditions:
Cnd $\to$ Itm at Area $\mid$ robot has Item $\mid$ robot at Area $\mid$ possible Stmt
    \end{lstlisting}
    \caption{Syntax of core language (subset). Reserved constants and variable
    names are marked in italic.}
    \label{fig:core-syntax}
  \end{figure}





  Intuitively, the core language interprets a program as a
  \emph{temporal goal} for the robot in a grid world.
  The model of the grid world consists of a tuple $(M, I, r)$ where
  $M$ is a two-dimensional grid of \emph{points} divided into free points and obstacles,
  $I$ is a set of \emph{items}, and $r$ is the robot.
  Each item $i \in I$ has associated attributes such as color, shape, and a unique identifier.
  Additionally, if it is not held by the robot, it has a position which is a point in $M$.
  The robot $r$ has a position in $M$ and holds a possibly empty set of items.

  The core language has a set of actions that manipulate the world and combines
  actions through standard imperative constructs such as sequencing, conditionals, or loops.
  We focus here on the non-standard parts of the language.

  The propositions of the logical language specify either sets of items or sets of
  points, and can be combined using Boolean operations as usual.
  The syntactic class ``Locations'' describes sets of points in the grid $M$.
  The user can, for instance, start with the entire grid $M$ using
  the keyword \lstinline{world} and filter the grid points to those which
  contain interesting items (e.g. one that is red) using the \lstinline{containing} clause:
  \begin{lstlisting}
  visit world containing item is red;
  pick item is red
  \end{lstlisting}
  Similarly, the syntactic class ``Items'' describes sets of items.
   The items of interest can be selected by (possibly Boolean combinations of) their attributes. Once selected, all such items can be picked or dropped by using the \lstinline{every} keyword
   \begin{lstlisting}
  foreach point in world containing item is red
      { visit point; pick every item is red }
  \end{lstlisting}

  \emph{Actions} modify the state of the world. For example, \lstinline{pick} changes the
  world from a state where there is an item $i$ in the current position of the
  robot to a world in which the item is being carried by the robot.
  Actions can be \emph{temporal}.
  The temporal action \lstinline{visit T}, for a set $T\subseteq M$ of points,
  requires that the robot is moved to some position in $T$.
  The temporal action \lstinline{visit $T$ while avoiding $A$} additionally requires
  that along the way the robot never visits any point in $A\subseteq M$:
  \begin{lstlisting}
  visit world containing item is red
      while avoiding world containing item is circle;
  pick item is color red
  \end{lstlisting}
  This can be written in linear temporal logic (LTL) as $\Diamond T$ and $\until{\lnot A}{T}$,
  respectively.\footnote{
	We remark that the temporal aspect of our core language is expressive enough to subsume LTL over finite traces with 
	propositions ranging over robot's locations (where \lstinline{visit while avoiding} 
and \lstinline{if move} constructs correspond to operators $\mathcal{U}$ and $\mathcal{X}$).
We omit the formal encoding.
In contrast to LTL-based robot planners \cite{fainekosTemporalLogicMotionPlanning,hadasLTLMop},
\tool's core language is closer to imperative programming languages likely to be more familiar
to programmers.
}

  In a complex command, only some part of a command may be realizable. Consider
  \begin{lstlisting}
  strict {while robot has item {drop item; move right}};
  \end{lstlisting}
  which instructs the robot to form, if possible, a horizontal line on the floor out of all the items
  the robot currently has (starting at robot's current position and to the right).
  The robot may be able to move right once but not as
  many times as it has items. The default behavior is lenient in that it
  executes those parts of the command which are realizable and prints a warning
  about those which are not. The \lstinline{strict} modifier allows to specify
  more rigid behavior that either performs the \emph{complete} action or no part
  of it.

  In a slight modification of the scenario above, if we want to place as many
  items to the right as possible:
  \begin{lstlisting}
  drop item;
  while possible {move right; drop item} {move right; drop item}
  \end{lstlisting}
  then we can use
  the \lstinline{while possible S T} construct which repeatedly executes $T$ while
  $S$ is realizable.

\subsection{Planning and User Interface}

  Given a goal expressed in the core language,
  \tool generates an execution plan for the robot to satisfy it in the grid world.
 The planner is based on A* search~\cite{Astar} with
  Christofides algorithm~\cite{christofides} for iterated reachability. 
  However,  for a different language, one could use a different planner
  \cite{hadasLTLMop,ankushDrona,antlab}. 
  Fast planning is crucial for user interaction and we found A* to be fast and sufficient to find plans.

  The generated plan is shown to the user visually in \tool's graphical
  interface (see \autoref{fig:simulationWorld} and our supplemental video) by dynamic simulation of the robot's moves in the UI. 
  This visual feedback is important to verify that a
  possibly complex command given to the robot works as expected. 
  Furthermore, ambiguities in the grammar can lead to several alternative plans being
  generated; the UI enables users to select the plan which best matches their intentions.
Finally, users can view a list of all induced grammar rules in a sidebar, alongside contexts in which a rule was defined.
They can additionally delete the rules defined by themselves.

\subsection{Usage Examples}

  Naturalization not only helps accommodate many different language styles, it can also significantly simplify programs.
  Consider the task of distributing items of different colors to different rooms.
  Conceptually, this is fairly simple: the robot should first gather all red items and put them into \lstinline{room1},
  then do the same for the blue items and \lstinline{room2} and so on.
  Figure \ref{fig:coreLanguageDefinitionOfSortingOnColors} shows an implementation of this specification in the core language.
  Each marked part corresponds to gathering items of a specific color 
  and bringing
  them to a specific room.
  Clearly, there is a significant amount of redundancy.
  \begin{figure}[t!]
    \begin{lstlisting}[frame=leftline,xleftmargin=15pt, xrightmargin=15pt]
  foreach point in world containing item has color red
    {visit point; pick every item is red };
  visit room1; drop every item is red;
    \end{lstlisting}
    \vspace{-5pt}
    \begin{lstlisting}[frame=leftline,xleftmargin=15pt, xrightmargin=15pt]
  foreach point in world containing item is green
    {visit point; pick every item is green};
  visit room2; drop every item is green;
    \end{lstlisting}
    \vspace{-5pt}
    \begin{lstlisting}[frame=leftline,xleftmargin=15pt, xrightmargin=15pt]
  foreach point in world containing item is blue
    {visit point; pick every item is blue};
  visit room3; drop every item is blue;
    \end{lstlisting}
    \vspace{-5pt}
    \begin{lstlisting}[frame=leftline,xleftmargin=15pt, xrightmargin=15pt]
  foreach point in world containing item is yellow
    {visit point; pick every item is yellow};
  visit room4; drop every item is yellow
    \end{lstlisting}
  \caption{Sorting items based on colors using the core language}
  \label{fig:coreLanguageDefinitionOfSortingOnColors}
  \end{figure}

  With naturalization, we can accomplish the same task by first defining
  \lstinline{gather red} as
    \begin{lstlisting}
  foreach point in world containing item is red
    {visit point; pick every item is red}
    \end{lstlisting}
  Then using this new command, we define \lstinline{red to room1} as
  \begin{lstlisting}
  gather red; visit room1; drop every item is red
    \end{lstlisting}
  and finally we can accomplish our complete task with
  \begin{lstlisting}
  red to room1; green to room2; blue to room3; yellow to room4
  \end{lstlisting}
  If we next want to put all items of different shapes to different rooms,
  the grammar induction allows us to re-use the commands defined above and simply
  write:
    \begin{lstlisting}
  triangle to room1; circle to room2; square to room3
    \end{lstlisting}


%

\subsection{Generalization}
\label{subsec:generalization}
We described how \tool expands its formal language by means of inducing new rules from users' definitions.
A potential issue with this approach is that users sometimes provide definitions which are
specific to the state of the presented world. 
Hence, the induced production
rules will also be specific and less usable in other scenarios.
For example, consider the situation from Figure~\ref{fig:simulationWorld} where 
a user wants to \emph{visit red triangle}. 
The user may provide the definition $y=$\lstinline{ move right}, since this command matches the user's 
expectation for the given world. 
However, in a different world, the desired item may not be right next to the robot and hence the induced rule will not be correct.

Even if the user provides a more general definition, it may still not work in a
different scenario. A user may define the utterance
\lstinline{pick 3 items} as $y=$\lstinline{ pick item; pick item; pick item}.
The naturalization method described in Section~\ref{sec:voxelurn} cannot produce
a meaningful production rule that would apply to commands of the form
\lstinline{pick $n$ items}, for arbitrary $n$.

We propose a solution in which Flipper synthesizes a number of definitions that
could be used instead of only the one supplied by user.
Let $w = (M, I, r)$ be the current state of the world.
We call two definitions $d_1$ and $d_2$ $w$-equivalent if their execution trace on the world $w$ is the same.
Since exploring the space of all possible $w$-equivalent definitions is not tractable, we limit the search to the following rewriting principles to produce a set $\mathcal{S}$ of $w$-equivalent definitions:
\begin{itemize}
	\item transforming a sequence of identical actions into a loop
	\item transforming the robot's movement actions into visiting a position defined by its coordinates or 
    by a predicate over items present at the field
	\item transforming the robot's picking or dropping actions into picking or dropping items defined by a predicate over items present at the field or at robot
\end{itemize} 
Once the set $\mathcal{S}$ is obtained, Flipper uses only the definition $d \in \mathcal{S} \cup \{y\}$ that maximizes the score function $\sigma$ to induce new rules. 

\paragraph*{Example}
In the world state depicted in Figure~\ref{fig:simulationWorld}, assume the user gave a too specific definition 
\lstinline{$y=$ move right} for the utterance \emph{visit red triangle}.
The set $\mathcal{S}$ contains (among others) the following alternative definitions: \lstinline{visit [4,0]}, \lstinline{visit world containing item}, \lstinline{visit world containing item is red}, \lstinline{visit world containing item is red and is triangle} out of which the last one scores the best and is selected to induce new grammar rules. Hence, from now on, \tool understands also e.g.\ \lstinline{visit blue circle}.

The scoring function $\sigma$ captures how suitable a definition is to the induction of new production rules and how well it matches the original user's intent expressed by the utterance.
Therefore, we define the score function as $\sigma(d, x, u) = (p_\theta(d | x, u),\textit{\text{sim}}(d,u))$ and compare different values of $\sigma$ lexicographically.
 $p_\theta$ (described in Section~\ref{sec:voxelurn}) captures the fit of the definition to the current semantic model. 
\textit{sim} takes a \textit{bag-of-words} representation of definition $d$ and utterance $u$.
After eliminating stop-words, it calculates the cosine similarity between the averages of word-embeddings~\cite{wordEmbeddings}.
The implementation uses pre-trained word vectors of dimension 100 from GloVe~\cite{pennington2014glove}.

Both the described generalization method and grammar induction described in Section~\ref{sec:voxelurn} can create incorrect rules. 
In case there are multiple parses (presumably, some of them incorrect), \tool offers a ranked list to the users.
The user then chooses the correct one, which updates the parameters $\theta$, effectively making it less likely for incorrect rules to be applied in the future.

\pgfplotsset{
    avgmin/.style={
        mark=-,
        error bars/.cd,
            y dir=minus,
            y explicit,
            error mark=-,
        /pgfplots/table/.cd,
            x=Group,
            y=Avg,
            y error expr=\thisrow{Avg} - \thisrow{Min}
    }
}

\pgfplotsset{
    avgmax/.style={
        mark=-,
        error bars/.cd,
            y dir=plus,
            y explicit,
            error mark=-,
        /pgfplots/table/.cd,
            x=Group,
            y=Avg,
            y error expr=\thisrow{Max} - \thisrow{Avg}
    }
}
\pgfplotsset{every error bar/.style={ultra thick}}

\pgfplotstableread{
Group Avg  Min Max
B-induced 54.58 19  110
}\datatableBInduced

\pgfplotstableread{
Group Avg  Min Max
C-induced 51.81 30  72
}\datatableCInduced

\pgfplotstableread{
Group Avg  Min Max
B-core 72.91 25 112
}\datatableBCore

\pgfplotstableread{
Group Avg  Min Max
C-core 33.5 17 54
}\datatableCCore

\newcommand{\plotTable}[1]{
\addplot+ [very thick, forget plot] 
  plot[very thick, error bars/.cd, y dir=plus, y explicit]
  table[x=Group,y=Avg,y error expr=\thisrow{Max}-\thisrow{Avg}] {#1};
\addplot+[very thick, xticklabels=\empty] 
  plot[very thick, error bars/.cd, y dir=minus, y explicit]
  table[x=Group,y=Avg,y error expr=\thisrow{Avg}-\thisrow{Min}] {#1};  
  }

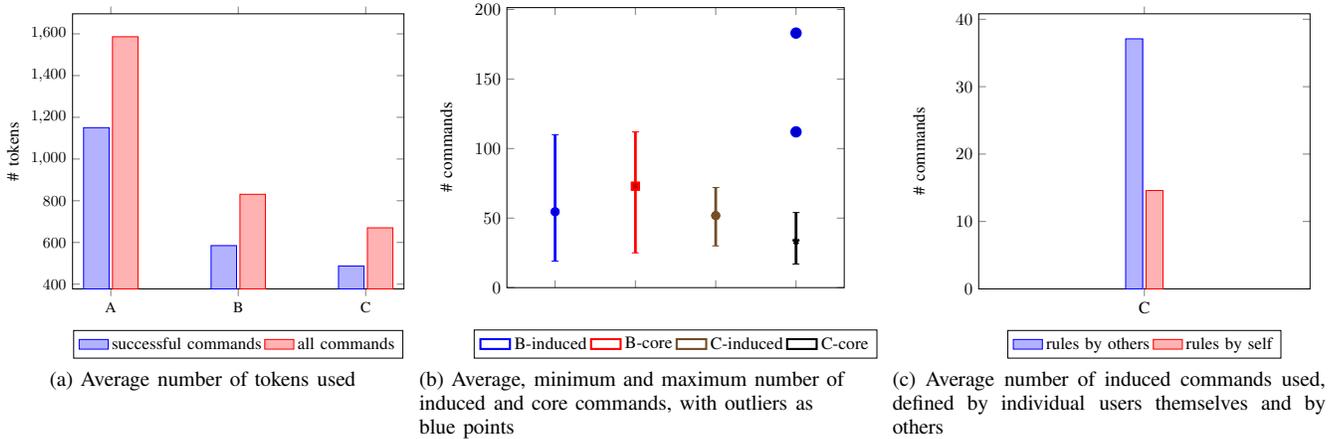
\begin{figure*}%
\centering
\subfloat[t][Average number of tokens used]{
  \resizebox{0.31\linewidth}{!}{
    \centering
        \begin{tikzpicture}
        \begin{axis}[
        ticklabel style = {font=\small},
            ybar,
            bar width = 15,
            area legend,
			enlarge x limits=0.15,
            legend style={at={(0.5,-0.15)},
              anchor=north,legend columns=-1},
            ylabel={\# tokens},
            symbolic x coords={A, B, C},
            xtick=data,
            ]
        \addplot coordinates {(A,1150.37 ) (B,585 ) (C,487.3) };
        \addplot coordinates {(A,1586.25 ) (B,831 ) (C,670.3) };
        \legend{successful commands, all commands}
        \end{axis}
        \end{tikzpicture}
  }
  \label{fig:numTokens}
}
\subfloat[Average number of induced and core commands, with maximum and minumum points labeled][Average, minimum and maximum number of\\ induced and core commands, with outliers as\\ blue points]{
  \resizebox{0.34\linewidth}{!}{
    \centering
        \begin{tikzpicture}
        \begin{axis}[
            bar width = 10,
            area legend,
			enlarge x limits=0.2,
,
            ymin = 0,
            legend style={at={(0.5,-0.15)},
              anchor=north,legend columns=-1},
            ylabel={\# commands},
            symbolic x coords={B-induced, B-core, C-induced, C-core},
            xtick={B-induced, B-core, C-induced, C-core},
            xticklabels={,,,}
            ]

\plotTable{\datatableBInduced}
\plotTable{\datatableBCore}
\plotTable{\datatableCInduced}
\plotTable{\datatableCCore}
\addplot+[only marks, forget plot, mark=*,mark size=3pt,,
        nodes near coords align={vertical},
        point meta=y,
        ]  coordinates {
    (C-core,183) (C-core,112)
};

  \legend{B-induced, B-core, C-induced, C-core}

        \end{axis}
        \end{tikzpicture}
  }
  \label{fig:groupBAndCCoreVsInduced}
}
\subfloat[t][Average number of induced commands used, defined by individual users themselves and by others]{
  \resizebox{0.31\linewidth}{!}{
    \centering
    \begin{tikzpicture}
      \begin{axis}[
          ybar,
          ymin = 0,
          area legend,
			enlarge x limits=0.5,
          legend style={at={(0.5,-0.15)},
            anchor=north,legend columns=-1},
          ylabel={\# commands},
          symbolic x coords={C},
          xtick=data,
          ]
        \addplot coordinates {(C,37.1)};
                \addplot coordinates {(C,14.6)};
        \legend{rules by others, rules by self}
      \end{axis}
    \end{tikzpicture}
  }
  \label{fig:groupCRulesInducedBySelfVsOthers}
}
\caption{Experimental results}
\label{3figs}
\end{figure*}

\section{Evaluation}


We performed an initial user study to evaluate whether actual users use 
naturalization and thus whether \tool makes programming robots easier.
Hence, we focused the evaluation on the following:
\begin{enumerate}
	\item \emph{usefulness of naturalization} as a comparison between the workload 
  of participants using \tool and those using only the core language
	\item \emph{usability of naturalization} in terms of the number of new specifications users	 defined and used
	\item \emph{naturalization across the group} in terms of the number of specifications
  used which were defined by others (without knowing them in advance).
\end{enumerate}

\subsection{Setup}
We created a list of 21 tasks, ranging from easy (e.g.
``get one green square'') to difficult (e.g. ``bring all red items to a
room that contains a yellow square''). The list of all tasks and
all our experimental data is available at \tool's website.
%
We recruited 31 participants, all with prior programming experience but without any
knowledge of the system. We split them into three groups: 
group $A$ was only able to use the core language;
group $B$ could additionally define new concepts;
and group $C$ could on top of that use the concepts defined by an expert user (familiar with the system and the language) as well as by other participants from the group.

At the start of the experiment, each group had 30 minutes to familiarize themselves with \tool and its core language by following a tutorial. 
Then the participants solved the 21 tasks as they saw fit, i.e., not necessarily in the most general way.
The average time needed to solve all tasks was 90 minutes (but no deadline was set). 

%

\subsection{Results}

\paragraph*{\textbf{Usefulness and Usability of Naturalization}}
To assess the usefulness of naturalization, we measure for each participant the
number of tokens used overall as well as in syntactically correct (parsable)
commands coming either from the core or the induced language.
The results are shown in Figure~\ref{fig:numTokens}.
Considering only parsable commands, members of group B on average had to use
50\% fewer tokens than the members of group A; and members of group C 
58\% fewer. This suggests that naturalization reduces users' effort.

Then, for participants from groups B and C we compare the number of induced
commands to the number of commands from the core language used to finish all
tasks.
The results in Figure~\ref{fig:groupBAndCCoreVsInduced} show that for groups $B$ and $C$,
a significant fraction of used commands were induced. 
Two participants in group  C, presumably with little programming experience, 
solved the tasks by explicit (step by step) instructions, with few definitions.
This resulted in a very large number of core commands (183 and 112, respectively).
Once these outliers were removed, we found that the average number of commands used was 33.5, ranging from 17 to 54.

Given that members of groups $B$ and $C$ were only motivated to use
naturalization to ease their way through the tasks, we conclude that the
naturalization technique was both useful and usable to them.

\paragraph*{\textbf{Naturalization across Users}}
Participants in group $C$ had access to the concepts defined by others. For this group,
Figure~\ref{fig:groupCRulesInducedBySelfVsOthers} shows how many of the induced
commands used were defined by the same participant or by others.

Participants were using a fair share of induced concepts, but still relied on
the core language (on average, 54 core language commands and 51 induced language
commands). When they used induced commands, they dominantly used commands defined
by others rather than commands defined by themselves.
Anecdotally, many participants used derived concepts without realizing they were derived.

Generalization (Section~\ref{subsec:generalization}) improved definitions in several cases.
For example, 
\lstinline{drop item; drop item; drop item} was rewritten \lstinline{repeat 3 times drop item} for 
the utterance \emph{drop 3 items} and  \lstinline{pick item is blue; pick item is blue} 
into \lstinline{pick every item is blue} for the utterance \emph{pick items is blue}.
 This is a first step towards completely eliminating a requirement that a subset of users has to master the core language.

%

\paragraph*{\textbf{Types of Defined Concepts}}
Upon closer inspection of the concepts the participants defined, we see that a majority falls into two categories:
\begin{inparaenum}[(1)]
\item simplifying individual commands and
\item defining functions.
\end{inparaenum}
Examples for the first case are
\lstinline{pick green square} defined as \lstinline{pick item is green and is square} and
\lstinline{visit empty space} defined as \lstinline$visit world minus {world containing item}$.

For the second case a simple example is
\lstinline{visit both triangle and green} defined as
\begin{lstlisting}
visit { {world containing item has shape triangle} and
  {world containing item has color green} }
\end{lstlisting}
There were also function definitions that involved previous function definitions,
such as \lstinline{line red} being defined as
\lstinline$fetch all red; while {robot has item} {drop item; move left}$.

\paragraph*{\textbf{Conclusion}}
The outcome of the user study is twofold. First, it shows naturalization is beneficial in defining complex temporal tasks, 
but only with some programming experience, as it allows users to define commands that generalize and can be reused.
Second, in order to serve non-programmers better, naturalization should be extended with orthogonal techniques which allow
users to express intent through explicit examples which can be generalized directly.

\section{Conclusion}

We have shown that naturalizing a domain-specific programming language is well suited
to provide a natural language interface to robot task specifications.
\tool provides the precision, expressivity, and extensibility of a programming
language, while ensuring a natural experience for humans.
\tool adapts its language to its users by learning new concepts from them. 
The results of our initial evaluation are encouraging and suggest
that a formal language for instructing robots can be turned with community
effort into a \emph{domain specific natural language}.


\balance





\bibliographystyle{IEEEtran}  
\bibliography{relatedWorkList}

\end{document}